\documentclass[preprint,12pt]{elsarticle}

\usepackage{amsmath,amssymb,amsfonts}
\usepackage{graphicx}
\usepackage{textcomp}
\usepackage{xcolor}
\usepackage{booktabs}
\usepackage{multirow}
\usepackage{array}
\usepackage{tabularx}
\usepackage{float}
\usepackage{url}
\usepackage[british]{babel}
\usepackage{csquotes}
\usepackage{microtype}
\usepackage{hyperref}
\hypersetup{
  pdftitle={Choosing a PEFT Variant for Per-Patient Dysarthric ASR: A Single-Speaker Case Study on Two ASR Bases},
  pdfauthor={Bernard Muller, Laszlo Toth, LaVonne Roberts}
}
\journal{Speech Communication}

\begin{document}

\begin{frontmatter}

\title{Choosing a PEFT Variant for Per-Patient Dysarthric ASR: A
Single-Speaker Case Study on Two ASR Bases}

\author[smf]{Bernard Muller\fnref{orcidBM}}
\fntext[orcidBM]{ORCID 0009-0004-7494-305X}
\author[usz]{L\'aszl\'o T\'oth\fnref{orcidLT}}
\fntext[orcidLT]{ORCID 0000-0003-0161-1375}
\author[smf]{LaVonne Roberts\corref{cor1}\fnref{orcidLR}}
\fntext[orcidLR]{ORCID 0009-0008-2883-3764}
\cortext[cor1]{Corresponding author}
\ead{lavonne@scottmorganfoundation.org}

\affiliation[smf]{organization={The Scott-Morgan Foundation},
  city={Torquay}, country={United Kingdom}}
\affiliation[usz]{organization={Institute of Informatics, University of Szeged},
  city={Szeged}, country={Hungary}}

\begin{abstract}
Per-patient adapters are the preferred production architecture for dysarthric
automatic speech recognition (ASR), yet parameter-efficient fine-tuning
(PEFT) variants have not been compared in the speaker-dependent, per-patient
regime. We present a single-speaker case study comparing seven LoRA-family
methods (LoRA, QLoRA, AdaLoRA, DoRA, LoHA, VeRA, VB-LoRA) on two production
bases (Whisper-large-v3 with Hungarian fine-tuning, and a multilingual
Qwen3-ASR-1.7B checkpoint) for one post-stroke Hungarian male speaker (S1,
409 utterances; severe dysarthria on auditory-perceptual clinical
assessment). Attention-projection adapters substantially improve CER on both
bases. Across three seeds, a paired bootstrap detects no significant
LoRA--DoRA difference ($p>0.5$; 13.86\,/\,13.90\,\% CER on Whisper,
28.10\,/\,28.33\,\% on Qwen3-ASR), so we adopt the simpler, cheaper LoRA.
Real 4-bit (NF4) QLoRA is worse on every seed and both bases
(14.56\,/\,30.09\,\% CER) with no memory saving at this scale, and LoHA,
VeRA, VB-LoRA and AdaLoRA do not reach the LoRA family, though LoHA still
gives an 18.6\,\% relative CER reduction on Whisper. On the same base, full
fine-tuning is more accurate (11.43\,\% CER), but a 115\,MB LoRA that also
adapts the feed-forward blocks reaches within 0.66\,pp of it at
$\approx$3.7\,\% of the per-patient storage. A 6-point enrollment grid shows
$\sim$5\,min of patient audio captures 45.6\,\% of the zero-shot-to-30-min
CER reduction, with further gains at 10 and 30\,min (caveat: one speaker,
one language, severe post-stroke dysarthria). Training scripts and recipes
will be released, source-available under a research-use licence, on
publication.
\end{abstract}


\end{frontmatter}

\section{Introduction}
\label{sec:intro}

Dysarthric speech recognition deployed for individual patients
increasingly relies on per-patient adapters: each user trains a small
adapter on their own voice, the shared base remains untouched, and
adapters are stored, versioned and shipped per patient. This is the production architecture
for clinical per-patient dysarthric-ASR deployments, and is
structurally similar to the patient-specific recipes reported by
Matsushima for Dutch dysarthric speakers
\citep{matsushima2022dutch}. The clinical motivation is unambiguous
--- patients want recognition of \emph{their} voice, not population
averages --- and the engineering consequence is that every megabyte of adapter and every minute of training is incurred once per patient, so total storage and compute scale linearly with the number of patients.

The choice of PEFT variant in this per-patient regime has not been
studied systematically for dysarthric speech. Dysarthric PEFT/adaptation
prior art does exist, but addresses different questions.
\citet{qi2023adapter} use adapter fusion with a Householder-transformation-based reparameterisation for target-speaker dysarthric adaptation.
\citet{wagner2025sap} compare full fine-tuning, plain LoRA
\citep{hu2022lora} and AdaLoRA \citep{zhang2023adalora} on the Speech
Accessibility Project corpus, with AdaLoRA being the strongest, in a
speaker-disjoint condition. The closest cross-variant comparison in
any clinical-adjacent population is \citet{ankita2026children} on
children's ASR, who report that LoHA \citep{hyeonwoo2023loha}
outperforms LoRA on Whisper-large-v3, particularly in the
speaker-disjoint condition. None of these match the production
per-patient case: speaker overlap is the defining property of
per-patient adaptation, and the question of which PEFT variant is best at the
small-data, speaker-dependent end of the spectrum has not been
answered. To our knowledge, this is the first speaker-dependent,
per-patient comparison of these seven LoRA-family variants on
dysarthric speech, under a fixed production recipe on the same
speaker and two ASR bases.

This paper answers that question on two production ASR bases for a
single real dysarthric speaker. Our contributions are:

\begin{enumerate}\setlength{\itemsep}{0pt}
  \item To our knowledge, the first speaker-dependent, per-patient
        cross-variant comparison of seven LoRA-family variants
        (LoRA, QLoRA, AdaLoRA, DoRA, LoHA, VeRA, VB-LoRA) on
        dysarthric speech.
  \item To our knowledge, the first per-architecture cross-comparison for
        dysarthric PEFT: same speaker, same data, same recipe, on a
        large encoder-decoder Transformer (Whisper-large-v3) and an
        LLM-decoder ASR (a multilingual production Qwen3-ASR-1.7B
        checkpoint).
  \item Two negative-result methodology findings published as
        named subsections: a warm-base regression analysis on Qwen3
        (Section~\ref{sec:qwen-regression}) and a follow-up dys-only pool
        ablation (Section~\ref{sec:dys-only}). Both pin the regression to
        narrow pretraining-mix, \emph{not} healthy-control-versus-dysarthric composition.
  \item A 6-point enrollment-time grid (Section~\ref{sec:enrollment-grid})
        on a production-equivalent recipe (DoRA $r{=}16$, which LoRA
        matches; Section~\ref{sec:multiseed}) showing that $\sim$5\,min of patient audio captures 45.6\,\% of the zero-shot-to-30-min CER reduction, with continued gains to 30\,min --- to our knowledge,
        the first such characterisation for LoRA-family PEFT on a fixed
        per-patient recipe.
  \item A multi-seed head-to-head with \emph{real} 4-bit QLoRA
        (Section~\ref{sec:multiseed}) showing LoRA and DoRA are statistically tied and 4-bit QLoRA lags behind both, and a released production
        recipe (LoRA $r=16$ on Whisper-large-v3 + Hungarian
        fine-tune; DoRA an alternative with no detected difference), source-available
        under a research-use licence, for per-patient adapter training.
  \item A same-base, same-data comparison and target-set attribution
        ladder (Section~\ref{sec:fullft-compare}) showing that encoder
        attention is the dominant LoRA lever and that extending the adapter
        to the feed-forward blocks reaches within 0.66\,pp CER of full
        fine-tuning at $\approx$3.7\,\% of its per-patient storage.
\end{enumerate}

\section{Background and prior art}
\label{sec:background}

\subsection{LoRA family taxonomy}
All seven variants we compare add trainable low-rank perturbations to
a frozen base. \textbf{LoRA} \citep{hu2022lora} factorises the weight update matrix as $\Delta W = BA$ with rank $r$. \textbf{QLoRA}
\citep{dettmers2023qlora} is defined by backpropagating the error through a
frozen 4-bit (NF4) quantised base into LoRA adapters, with double
quantisation and paged optimisers; its purpose is memory reduction,
not a change of the adapter parameterisation. \textbf{AdaLoRA}
\citep{zhang2023adalora} reallocates a parameter budget across layers
using importance scores. \textbf{DoRA} \citep{liu2024dora} decomposes
the update into magnitude and direction components, with magnitude
trained directly. \textbf{LoHA} --- the PEFT library's Hadamard-product adapter, derived
from the \emph{FedPara} parameterisation \citep{hyeonwoo2023loha} (a
federated-learning antecedent, not a paper titled LoHA) --- uses a
Hadamard product of two low-rank pairs for effective rank $\sim r^2$
at the same storage cost. \textbf{VeRA} \citep{kopiczko2024vera}
freezes shared random projections and trains only per-layer
scaling vectors. \textbf{VB-LoRA} \citep{li2024vblora} shares a
vector bank across layers with per-layer mixing weights.

\subsection{Dysarthric ASR adaptation prior art}
\citet{matsushima2022dutch} reports substantial WER reductions on
Dutch dysarthric speech from speaker-dependent re-fine-tuning of a
self-supervised model with only $\sim$10\,min of target-speaker
audio, establishing per-patient adaptation as a strong single
intervention. \citet{qi2023adapter} adapt to a
target dysarthric speaker via adapter fusion and a Householder-transformation-based reparameterisation, matching a baseline at $\sim$1/3 the parameters.
\citet{wagner2025sap} compare full fine-tuning, LoRA and AdaLoRA on
Whisper-large-v3 for the Speech Accessibility Project --- the closest
contemporary dysarthric-PEFT comparison, with AdaLoRA outperforming both other methods, but in a speaker-disjoint condition.
\citet{ankita2026children} compare LoRA variants on Whisper for
children's ASR and report LoHA as the winner, motivating the
present study. Beyond LoRA-family PEFT, speaker adaptation for
disordered speech has been pursued through prompt-tuning
(\citet{jiang2024perceiver} attach a trainable Perceiver speaker
prompt to Whisper for Chinese disordered speech) and through
non-PEFT routes such as voice/rhythm conversion
\citep{elhajal2025rnv}; \citet{mihajlik2025tts} report a personalised
Hungarian dysarthric speech-to-text case study via TTS-driven data
augmentation.
Personalisation of ASR for atypical speech from a small amount of per-speaker data has
a substantial track record: \citet{shor2019personalizing} personalise
recognisers for dysarthric and accented speakers from a few minutes of
data, \citet{tobin2022personalized} characterise how per-speaker accuracy
scales with disordered-speech dataset size, and \citet{tomanek2021residual}
adapt foundation ASR to atypical and accented speech with residual
adapters --- the closest parameter-efficient precedent to the present work,
though without a LoRA-family variant comparison. Parameter-efficient
personalisation within the LoRA family is now emerging:
\citet{joseph2024speaker} and \citet{baby2024robust} apply DoRA and a
generalised low-rank adaptation to speaker personalisation, and
\citet{pokel2025variational} propose a variational low-rank adapter for
personalised impaired-speech recognition, while \citet{huber2026adapting}
report a single-speaker case study adapting foundation ASR models to
dysarthric speech.
None of these answer the per-patient
\emph{speaker-dependent} PEFT-variant question we study: which
low-rank adapter to default to under a fixed sparse-data recipe.

\subsection{FiLM speaker-conditioning: the closest alternative
mechanism}
\label{sec:film-prior}

The most relevant contemporary alternative-mechanism prior art is the work of \citet{lopez2026film}, who replace per-patient LoRA with feature-wise
linear modulation (FiLM)~\citep{perez2018film} conditioned on a speaker x-vector. They
target Voxtral-Mini, a SpeechLLM combining a Whisper-large-v3 audio
encoder with a Ministral-3B language decoder.
A SiAmResNet34 network extracts per-speaker x-vectors and a small projection
head emits per-layer $(\gamma, \beta)$ modulation parameters
injected into the frozen base; this adds 73.5\,M trainable parameters
($\sim$1.6\,\% of Voxtral-Mini's 4.7\,B). On the TORGO and NeuroVoz databases they
report two findings that are in tension. As regards ASR, an F-LoRA hybrid is strongest
on NeuroVoz (4.07\,\% WER) and competitive on TORGO (12.71\,\% WER,
behind full fine-tuning's 10.97\,\%); FiLM-only speaker conditioning
trades several pp of WER for its other advantage. As regards the model's multi-task instruction-following
capability --- measured on a SpeechLLM MCQA benchmark --- plain LoRA
on a per-patient pool sharply reduces MCQA accuracy from
52.7\,\% (frozen base) to 8.6\,\%, while FiLM preserves the base's
MCQA accuracy by design (the frozen weights are never updated).
This is also the framing's fleet-storage argument: each enrolled
patient costs only $\sim$2\,KB (an x-vector) plus shared FiLM
projection heads, against $\sim$60\,MB of per-patient LoRA adapter
weights at the rank we use here.

We treat FiLM as the closest contemporary alternative mechanism, not
as a configuration, for three reasons. (a)~FiLM's premise --- distinguishing
patients via x-vectors --- requires $N>1$ speakers; on a single
training speaker the x-vector is constant and the FiLM head
degenerates to a per-layer fixed bias, weaker than LoRA on
the same layers. (b)~Voxtral-Mini does not list Hungarian among its
supported languages, ruling it out for the same pretraining-mix
narrowness reason that emerges in our NeMo backbone
Section~\ref{sec:nemo-footnote}. (c)~Our controlled-variable design holds
base + data + recipe constant and varies only the PEFT variant,
which keeps the comparison interpretable within the LoRA family. A
proper FiLM-vs-DoRA benchmark would require a fleet-level multi-patient
comparison regarding storage-per-patient, CER, and multi-task retention;
we defer it explicitly to future work (Section~\ref{sec:conclusion}).

\section{Method}
\label{sec:method}

\subsection{Data}
We use a single post-stroke Hungarian male speaker (S1). The participant's
clinical speech-language-therapy records from early 2023 (about two years before the corpus itself was recorded, in early 2025) characterise the
dysarthria as severe, on auditory-perceptual assessment and without a
standardized severity instrument; this is consistent with the severe
characterisation reported by \citet{mihajlik2025tts} for the same speaker.
No standardized, independent severity rating of the corpus recordings is
available, so severity here is a descriptive clinical characterisation
rather than a validated instrument-based classification --- we use the
categorical terminology surveyed by \citet{stipancic2021severity} only as
descriptive vocabulary. The
corpus comprises 409 utterances (55\,min)
under a signed Data Sharing Agreement that allows research use but
forbids redistribution of the audio itself (see
Section~\ref{sec:practical-recipe}). Splits are speaker-internal by content
type: 262 training utterances (195 read-sentence + 67 narrative),
40 validation, and 107 evaluation. The training pool breaks down as
21.12\,min read sentences + 11.56\,min narrative (32.68\,min total).

S1 is the same participant, and the corpus the same, as in Mihajlik et al.\
\citep{mihajlik2025tts}: the 107-utterance evaluation split, the
40-utterance validation split, and the 195 read-sentence training items are
identical across the two studies (our per-patient training pool
additionally includes 67 narrative utterances, giving 262). The two studies
ask different questions --- Mihajlik et al.\ compare full fine-tuning of
several backbones with TTS-driven data augmentation, whereas we compare
parameter-efficient adapter variants on a fixed per-patient recipe --- but
because they share a participant and a corpus, we state the overlap
explicitly rather than leave it to citation.

\subsection{Bases}
\begin{itemize}\setlength{\itemsep}{0pt}
  \item \textbf{Whisper-large-v3} \citep{radford2023whisper}: 1.55\,B
        encoder-decoder Transformer. We use a \textbf{merged HU\,FT}
        variant: Whisper-large-v3 fine-tuned on the 38K-utterance
        Hungarian pool described in Section~\ref{sec:qwen-regression},
        with full-weight merge into the base. Zero-shot on the S1
        test split after merge results in 29.46\,\% CER.
  \item \textbf{Qwen3-ASR-1.7B}: an \emph{internal} multilingual
        production checkpoint --- a dysarthric-pool fine-tune of the
        public 1.7\,B base \citep{alibaba2024qwen3asr}, not itself a
        public release. It is a 1.7\,B AuT encoder + Qwen LLM decoder
        trained on a 10-language pool including TORGO, UASPEECH,
        EasyCall and other dysarthric corpora. On our internal
        multilingual evaluation (131{,}849 utterances across 10 languages)
        it reaches 2.91\,\% CER / 5.49\,\% WER. Zero-shot on the S1
        test split gives 49.46\,\% CER. Because this base is not public,
        results on it are reproducible in recipe but not from publicly available files alone (Section~\ref{sec:practical-recipe}).
\end{itemize}

\subsection{PEFT variants}
\label{sec:peft-variants}
All seven variants are applied via the HuggingFace
\textsc{PEFT} library \citep{peft2023library} on top of
\textsc{transformers} \citep{wolf2020transformers}, with the following parameters:
LoRA $r{=}16,\alpha{=}32$;
DoRA $r{=}16$;
QLoRA $r{=}16$ (real 4-bit NF4 frozen base with double quantisation,
in the multi-seed head-to-head; the seven-variant comparison of Table~\ref{tab:main} additionally lists a QLoRA variant with 4-bit quantisation disabled as a LoRA-equivalent control);
LoHA $r{=}8$ (effective rank $\sim$64);
AdaLoRA init\_$r{=}24\to$ target\_$r{=}16$;
VeRA $r{=}256$;
VB-LoRA $r{=}16$, vector\_bank $=256$.
Target modules are the attention projections, matched by module name
across the whole model. For Whisper this selects the $q$/$k$/$v$/\texttt{out\_proj}
in the encoder self-attention and the decoder self- and cross-attention
(384 modules; no feed-forward layers). For Qwen3-ASR it selects the
LLM-decoder attention ($q$/$k$/$v$/$o$) together with audio-encoder
projection and convolution outputs (\texttt{out\_proj}, \texttt{conv\_out},
\texttt{proj1}, \texttt{proj2}). Both bases therefore adapt encoder and
decoder attention while leaving the feed-forward blocks and the remaining
encoder weights frozen. We report the trainable-parameter count, adapter
size and adapted-module inventory per variant so the adaptation surface is
explicit, and Section~\ref{sec:fullft-compare} isolates the contribution of
each target group (encoder attention, decoder attention, feed-forward)
directly. The shared base --- including the bulk of the audio encoder ---
is cached once and reused across patients; only the low-rank adapter is
stored per patient.

\subsection{Training recipe}
The training recipe was identical across all configurations: AdamW (weight decay 0), learning rate
$1\!\times\!10^{-4}$, bf16 (a 16-bit floating-point format), batch 4 $\times$ grad-acc 4, a fixed
5 epochs ($\sim$80 optimiser steps), cosine schedule with 10\,\%
warmup, seed 42 (42/43/44 for the multi-seed runs). Training is
fixed-budget --- no early stopping and no intermediate checkpointing;
the final-epoch model is evaluated. The 40-utterance validation split
is held out but was not used for checkpoint selection in these runs.
We varied one factor: the PEFT variant.

\subsection{Hardware and software}
\label{sec:hardware}
We employed a DGX Spark (GB10, sm\_121) in NGC container
\url{nvcr.io/nvidia/pytorch:25.10-py3}
(PyTorch 2.9.0a0, CUDA 13.0). We used \textsc{transformers} 4.51.x and
\textsc{PEFT} 0.19.1 (the latter required for VeRA and VB-LoRA
support); \textsc{bitsandbytes} 0.49.2 for the 4-bit NF4 QLoRA configurations.
The released repository pins these exact versions.

\subsection{Evaluation}
The 107-utterance test split shares no reference text with the 262-utterance
training pool or the 40-utterance validation split (none of the 107 evaluation utterances shares its reference text with either split, after Unicode NFC normalisation), and S1's data was excluded from the Hungarian
fine-tuning pool used to build the warm base
(Section~\ref{sec:qwen-regression}); the multilingual Qwen3-ASR checkpoint
contains no S1 data. The split is therefore text-disjoint. This same
107-utterance set was, however, used by \citet{mihajlik2025tts} and ---
within this study --- across the seven-variant screening, the contender
selection and the multi-seed head-to-head; we therefore treat it as a
reused evaluation benchmark rather than an independently held-out set, and a
fresh, session-disjoint confirmatory set is deferred to future work
(Section~\ref{sec:conclusion}).
We applied greedy decoding on the 107-utterance test split per configuration. CER and WER are calculated from a single-reference --- the corpus ships with one reference
transcription per utterance --- and all reported numbers in this
paper are greedy single-reference results. Primary metric is CER;
WER is secondary. We additionally report trainable parameter count,
serialised adapter size (MB) and wall-clock training time (min).
CER/WER are computed after Unicode NFC normalisation, lower-casing,
and restriction to the Hungarian alphabet (a--z, the accented set
\textit{á é í ó ö ő ú ü ű}, and space; all other characters collapse
to a single space); no punctuation, casing, or numeral expansion is
applied beyond this. The peak VRAM usage was estimated by \texttt{torch.cuda.max\_memory\_allocated()} read after
\texttt{reset\_peak\_memory\_stats()} at load.

\subsection{Reproducibility}
Decoding uses greedy search with the Hungarian language token, the
transcribe task, and \texttt{max\_new\_tokens}\,=\,200. The adapted
modules are exactly those in Section~\ref{sec:peft-variants}: the attention
projections across encoder and decoder (no feed-forward layers for the
Whisper cells), with the feed-forward blocks and the remaining encoder
weights frozen; all other hyperparameters are as given in the fixed recipe
above. Training scripts,
per-variant configs, and the Pareto / enrollment runners are released
source-available under a research-use licence (\texttt{per-patient-peft};
tagged release on publication). The S1 corpus, trained adapters, and
per-utterance JSON outputs remain restricted under the data-sharing
agreement (Section~\ref{sec:practical-recipe}), so the main results are reproducible in code logic but not from publicly available files alone.

\subsection{Use of AI tools in the research process}
\label{sec:ai-research}
The first author works through an AI-assisted eye-gaze interface and used
Claude Opus 5 (Anthropic) throughout this study as a research tool: to draft and
debug the training, evaluation, and plotting scripts released with this
paper; to compute the CER/WER, bootstrap, and summary statistics reported in
Section~\ref{sec:results}; and to generate Figures~\ref{fig:pareto}
and~\ref{fig:enrollment} from the underlying result files via the released
plotting scripts. All code was executed by the authors on the hardware in
Section~\ref{sec:hardware}, every reported number was read by an author from
the resulting output files, and all experimental design decisions,
interpretations, and conclusions are the authors' own. The released
repository contains the scripts as run, so every figure and table can be
regenerated from the result files without the AI tool.

\section{Results}
\label{sec:results}

\subsection{Main comparison}
\begin{figure}[t]
\centering
\includegraphics[width=\linewidth]{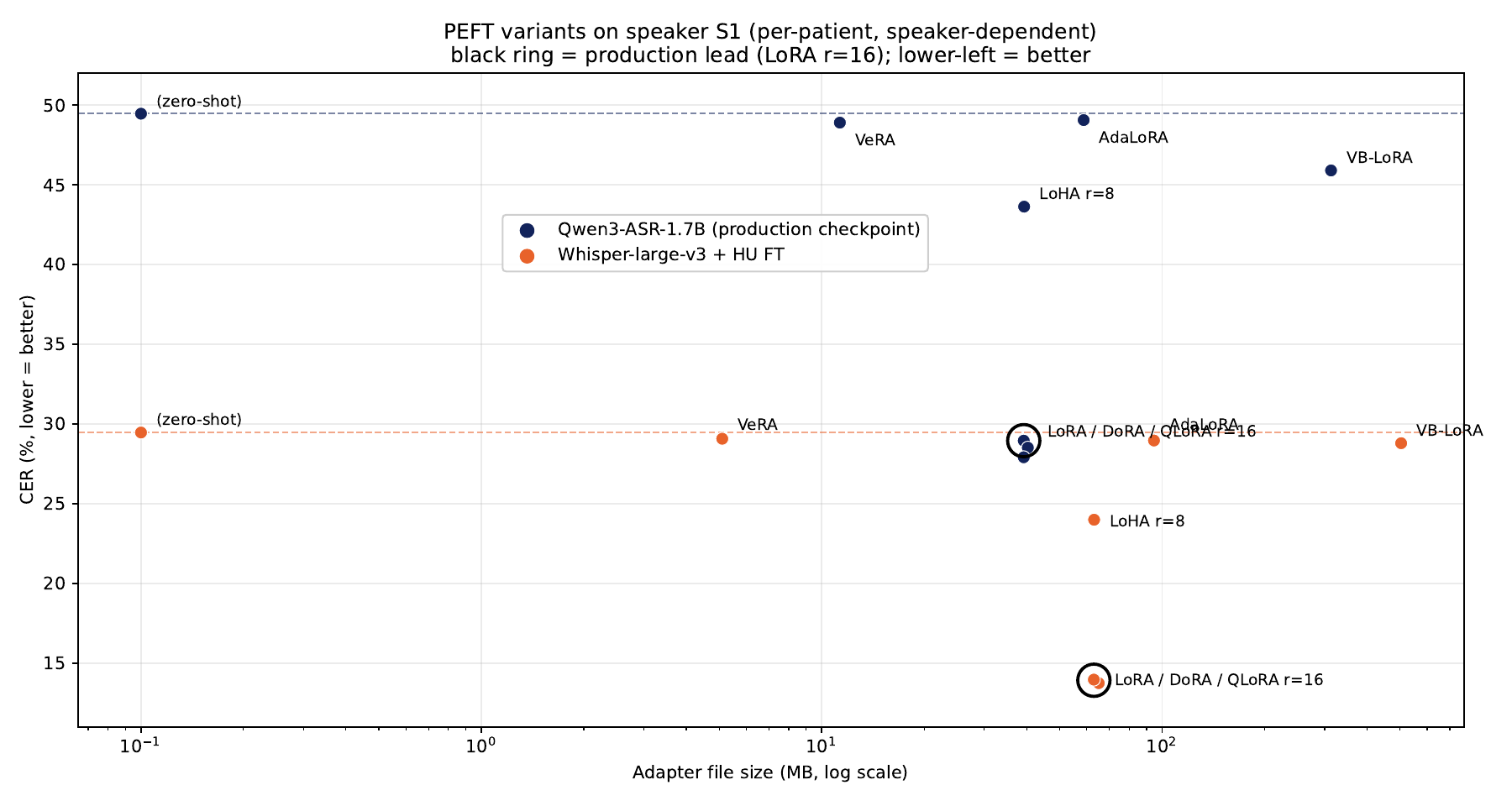}
\caption{Pareto chart of CER vs adapter size (MB, log axis) for all
16 configurations. Whisper-large-v3 + HU\,FT configurations are orange circles; Qwen3-ASR
production-checkpoint configurations are blue circles. Dashed lines mark the
per-base zero-shot CER. The tied top variants LoRA and DoRA
($r{=}16$) sit on the Pareto front for both bases.}
\label{fig:pareto}
\end{figure}

Table~\ref{tab:main} reports CER, WER, trainable parameter count,
adapter size and wall-clock training time for all 16 (2$\times$8) configurations, and
Figure~\ref{fig:pareto} plots CER against adapter size for the same
configurations. All
CER/WER values are greedy single-reference on the held-out test
split.

\begin{table}[t]\centering\scriptsize
\caption{Variant comparison: seven PEFT variants on two production bases,
single dysarthric speaker (S1, Hungarian), 107-utterance held-out
evaluation, single seed. $\Delta$CER is relative improvement vs the per-base zero-shot baseline. This table compares all seven variants, separating
the contending strict-sense LoRA family (LoRA, DoRA, QLoRA) from the rest; the
$^{\dagger}$QLoRA row here uses the QLoRA \emph{parameterisation
without} 4-bit quantisation (a LoRA-equivalent control). The
within-family winner is decided not here but in the multi-seed
head-to-head with \emph{real} 4-bit QLoRA (Table~\ref{tab:multiseed}),
which shows LoRA and DoRA are statistically tied and 4-bit QLoRA lags behind both. We therefore do not bold a single-seed winner.}
\label{tab:main}
\setlength{\tabcolsep}{2pt}
\begin{tabular*}{\linewidth}{@{\extracolsep{\fill}}lrrrrrr@{}}
\toprule
Variant & CER (\%) & WER (\%) & $\Delta$CER (rel \%)
& Params & Adapter (MB) & Train (min)\\
\midrule
\multicolumn{7}{@{}l}{\textit{Whisper-large-v3 + HU\,FT}}\\
(zero-shot)   & 29.46 & 49.53 & 0.0    & 0           & 0     & --\\
DoRA $r{=}16$ & 13.74 & 29.50 & $+$53.4
& 16{,}220{,}160 & 65.1 & 12.7\\
LoRA $r{=}16$         & 13.92 & 29.36 & $+$52.8 & 15{,}728{,}640 & 63.0 & 6.7\\
QLoRA$^{\dagger}$ $r{=}16$ & 13.96 & 29.86 & $+$52.6 & 15{,}728{,}640 & 63.0 & 6.7\\
LoHA $r{=}8$          & 23.99 & 45.05 & $+$18.6 & 15{,}728{,}640 & 63.1 & 8.0\\
VB-LoRA               & 28.79 & 49.02 & $+$\phantom{0}2.3  & 125{,}837{,}312 & 503.5 & 8.1\\
AdaLoRA               & 28.96 & 49.10 & $+$\phantom{0}1.7  & 23{,}602{,}176  & 94.6  & 7.5\\
VeRA                  & 29.07 & 48.88 & $+$\phantom{0}1.3  & 589{,}824       & 5.1   & 9.0\\
\addlinespace
\multicolumn{7}{@{}l}{\textit{Qwen3-ASR-1.7B (multilingual production checkpoint)}}\\
(zero-shot)   & 49.46 & 76.86 & 0.0    & 0           & 0     & --\\
QLoRA$^{\dagger}$ $r{=}16$ & 27.91 & 52.35 & $+$43.6
& 9{,}789{,}440  & 39.2 & 2.2\\
DoRA $r{=}16$ & 28.51 & 53.65 & $+$42.4 & 10{,}063{,}872 & 40.3 & 3.7\\
LoRA $r{=}16$ & 28.95 & 52.78 & $+$41.5 & 9{,}789{,}440  & 39.2 & 2.2\\
LoHA $r{=}8$  & 43.63 & 72.23 & $+$11.8 & 9{,}789{,}440  & 39.3 & 2.9\\
VB-LoRA       & 45.90 & 73.46 & $+$\phantom{0}7.2 & 78{,}323{,}712 & 313.4 & 3.0\\
VeRA          & 48.90 & 77.37 & $+$\phantom{0}1.1 & 328{,}448       & 11.3  & 2.4\\
AdaLoRA       & 49.06 & 76.28 & $+$\phantom{0}0.8 & 14{,}689{,}224  & 58.8  & 2.6\\
\bottomrule
\end{tabular*}
\end{table}

\subsection{Multi-seed head-to-head with real 4-bit QLoRA}
\label{sec:multiseed}
The comparison of Table~\ref{tab:main} isolates the LoRA family (LoRA,
DoRA, QLoRA) as the only variants that adapt usefully on either base.
To decide \emph{within} that family --- and to test QLoRA as it is
actually defined --- we re-ran LoRA, DoRA and \emph{real} 4-bit NF4
QLoRA over three seeds (42/43/44) on both bases, with per-utterance
hypotheses logged for paired significance testing
(Table~\ref{tab:multiseed}). Two findings emerged.
\textbf{(i) No significant LoRA--DoRA difference was detected.} Their
three-seed means differed by $\le$0.23\,pp CER on either base, well
inside the spread caused by varying the seed, and a 10,000-resample paired bootstrap on the pooled per-utterance
hypotheses across the three seeds (deltas computed from unrounded values;
resampler seed 0) detected no significant difference (Whisper
$\Delta$CER $+0.03$, 95\%\,CI $[-0.17,+0.25]$, $p{=}0.79$; Qwen3
$\Delta$CER $+0.23$, CI $[-0.50,+0.96]$, $p{=}0.55$). DoRA's apparent
single-seed lead on Whisper in Table~\ref{tab:main} did not survive re-seeding --- LoRA's three-seed mean was in fact marginally lower on
both bases. We read this as a practical tie rather than proven
equivalence: we did not pre-register an equivalence margin, and the
intervals are utterance-level on one speaker's test split, so they
bound system-level uncertainty here, not patient-population
variability. \textbf{(ii) Real 4-bit QLoRA was consistently worse.} It lost to LoRA on every seed and both bases ($+$0.69\,pp CER on
Whisper, $+$1.99\,pp on Qwen3), and --- at this 1.5--1.7\,billion-parameter scale --- yielded no peak-VRAM saving (16.9 vs 14.8\,GiB on Whisper;
parity on Qwen3). In this small-data per-patient regime the
4-bit base traded accuracy for a memory benefit that did not materialise. The $\sim$2$\times$ absolute gap between the two 4-bit QLoRA CERs (14.56 vs 30.09\,\%) is not quantisation-specific: it tracks the same $\sim$2$\times$ gap seen for LoRA and DoRA (13.86 vs 28.10\,\%) and reflects Qwen3-ASR's higher zero-shot CER on this speaker (49.46 vs 29.46\,\%), not a quantisation$\times$architecture interaction; the QLoRA penalty \emph{relative to} LoRA is comparable on both bases ($+$0.69\,pp Whisper, $+$1.99\,pp Qwen3).

\begin{table}[t]\centering\scriptsize
\caption{Multi-seed head-to-head (3 seeds 42/43/44, mean\,$\pm$\,sd)
for the three contending variants, with \emph{real} 4-bit NF4 QLoRA.
LoRA and DoRA are statistically tied on both bases; 4-bit QLoRA is
behind both for every seed.}
\label{tab:multiseed}
\begin{tabular}{llrr}
\toprule
Base & Variant & CER (\%) & WER (\%)\\
\midrule
\multirow{3}{*}{Whisper-large-v3 + HU\,FT}
& LoRA $r{=}16$         & 13.86 $\pm$ 0.07 & 29.24 $\pm$ 0.18\\
& DoRA $r{=}16$         & 13.90 $\pm$ 0.07 & 29.26 $\pm$ 0.07\\
& QLoRA $r{=}16$ (4-bit) & 14.56 $\pm$ 0.07 & 30.10 $\pm$ 0.36\\
\midrule
\multirow{3}{*}{Qwen3-ASR-1.7B (prod.)}
& LoRA $r{=}16$         & 28.10 $\pm$ 0.60 & 52.28 $\pm$ 0.45\\
& DoRA $r{=}16$         & 28.33 $\pm$ 0.30 & 52.86 $\pm$ 0.46\\
& QLoRA $r{=}16$ (4-bit) & 30.09 $\pm$ 0.34 & 56.42 $\pm$ 0.35\\
\bottomrule
\end{tabular}
\end{table}

\subsection{Cross-base variant ranking}
Table~\ref{tab:cross-base} reports the comparison ranking by CER for each
base. The LoRA family dominates on both bases; the storage-efficient
(VeRA, VB-LoRA) and budget-reallocation (AdaLoRA) variants
under-perform on both. Within the LoRA family the comparison's single-seed ordering is unreliable (Table~\ref{tab:multiseed}): LoRA and DoRA are
tied on re-seeding and real 4-bit QLoRA falls behind both. We
therefore adopt LoRA as the per-patient production lead on grounds of
simplicity and cost (half the Whisper training time of DoRA, no
magnitude-vector overhead), with DoRA an alternative with no detected difference (Section~\ref{sec:discussion}, finding 2).

\begin{table}[t]\centering\scriptsize
\caption{Single-seed \emph{comparison} ranking by CER per base (from
Table~\ref{tab:main}; QLoRA here is the non-quantized control). The LoRA
family separates cleanly from the rest, but the within-family order
is not reliable --- on re-seeding with real 4-bit QLoRA, LoRA and
DoRA tie and 4-bit QLoRA drops behind both (Table~\ref{tab:multiseed}).}
\label{tab:cross-base}
\begin{tabular}{lrr}
\toprule
Variant & Whisper rank & Qwen3 rank\\
\midrule
DoRA $r{=}16$  & 1 & 2\\
LoRA $r{=}16$  & 2 & 3\\
QLoRA$^{\dagger}$ $r{=}16$ (no 4-bit) & 3 & 1\\
LoHA $r{=}8$   & 4 & 4\\
VB-LoRA        & 5 & 5\\
AdaLoRA        & 6 & 7\\
VeRA           & 7 & 6\\
\bottomrule
\end{tabular}
\end{table}

\subsection{NeMo-backbone negative result}
\label{sec:nemo-footnote}

We additionally evaluated two NeMo Granary-pretrained backbones
(NVIDIA Parakeet-TDT-0.6B-v3 \citep{nvidia2025parakeet} and
Canary-1B-v2 \citep{nvidia2025canary}; both licensed under CC-BY-4.0, both natively
Hungarian-capable) under simple LoRA fine-tuning on S1's 262-utterance
training pool. Parakeet regresses relative to zero-shot and Canary collapses outright:
Parakeet-TDT-v3 + LoRA $r{=}16$ on the encoder degrades from
50.59\,\% to 69.07\,\% CER (+18.48\,pp; the encoder LoRA broke the
RNN-T joint network's Hungarian prior, producing English/Russian
fragment hallucinations); Canary-1B-v2 + LoRA $r{=}16$ on encoder +
transformer decoder collapses to 191.22\,\% CER, with seq2seq EOS collapse (the decoder fails to emit the end-of-sequence token, so generation does not terminate) producing repetition loops (``\textit{a szoba a szoba a
szoba\ldots}'') until reaching the \verb|max_new_tokens| limit.

Under the simple LoRA recipes tested, both NeMo backbones regress
rather than improve. The tested recipes are aligned with the
Whisper and Qwen3 configurations of Section~\ref{sec:results}: encoder-only
($r{=}16$) for the RNN-T Parakeet, encoder + transformer-decoder
($r{=}16$) for the seq2seq Canary, with no per-backbone hyperparameter
tuning. The most parsimonious interpretation is that adapter
fine-tuning on a model whose pretraining mix has too narrow a
Hungarian or atypical-speech buffer rotates the small available
adaptation material enough to \emph{break} the existing language
prior --- and that Whisper and the Qwen3-ASR production checkpoint
survive because their broader pretraining mixes leave room to move more efficiently. We deliberately do not claim the failure to be structural
in a stronger sense: we have not run deeper LoRA targets (e.g. the RNN-T joint network on Parakeet, cross-attention-only on Canary) or
higher-rank LoHA / DoRA variants on these backbones, and we cannot
rule out that one of those recipes would recover the regression.
The negative result here therefore supports the narrow-pretraining
hypothesis under simple LoRA recipes; a definitive per-aetiology
multi-LoRA-recipe sweep on NeMo backbones is queued for a follow-up
study (Section~\ref{sec:conclusion}). The integration walls hit when
getting NeMo 2.x backbones into a per-patient PEFT recipe (Lightning
package mismatch, Lhotse strict-int schema, transformer-decoder
linear-layer naming) are documented in the released scripts.

\subsection{Picking the warm base: Qwen3 HU-FT regression}
\label{sec:qwen-regression}

A natural design for cross-base comparison is to construct an
apples-to-apples Hungarian warm base for both Whisper-large-v3 and
Qwen3-ASR-1.7B by fine-tuning each on the same 38K-utterance
Hungarian pool (Common Voice + FLEURS + Hungarian\_Dysarthria +
VoxPopuli\_hu, S1 excluded). The Whisper chain transferred cleanly
(zero-shot $\to$ HU\,FT $\to$ S1 LoRA = 32.2\,\% $\to$ 29.46\,\%
$\to$ 13.9\,\% CER on the 107-utterance test split). The Qwen3 chain did not.

The first signal was an initial fine-tune whose final-epoch evaluation
reversed the direction of effect: dysarthric CER on S1 rose after the same
Hungarian fine-tune that had helped Whisper, rather than falling. We
initially abandoned the Qwen3-1.7B + HU\,FT path, then investigated it
properly under a single controlled run to turn that preliminary observation
into an identifiable per-checkpoint trajectory. All Qwen3 HU\,FT figures we
report come from this controlled rerun (Table~\ref{tab:qwen-regression}),
not from the abandoned preliminary run.
For diagnostic purposes, we re-ran the Qwen3 HU\,FT with intermediate
checkpoints (saved after 25\,\%, 50\,\%, 75\,\% and 100\,\% of the optimiser steps in one epoch) and
evaluated each against both the S1 test split (107 dysarthric utterances) and a
50-utterance CV\_Hungarian sample (see Table~\ref{tab:qwen-regression}):

\begin{table}[t]\centering\scriptsize
\caption{Qwen3-ASR-1.7B Hungarian fine-tune trajectory.
Clean-Hungarian CER (on a set that is itself inside the fine-tuning pool)
holds while dysarthric CER degrades; Section~\ref{sec:dys-only} shows this
preservation is pool-dependent, not selective robustness.}
\label{tab:qwen-regression}
\begin{tabular}{lrr}
\toprule
Checkpoint & S1 test CER & CV\_Hungarian CER\\
\midrule
Base zero-shot   & 43.61\,\%          & \textbf{7.19\,\%}\\
25\,\% (step 602)  & 54.56\,\%        & 9.43\,\%\\
50\,\% (step 1204) & 53.99\,\%        & 8.06\,\%\\
75\,\% (step 1806) & 60.45\,\%        & 7.31\,\%\\
100\,\% (step 2408) & \textbf{67.16\,\%} (+23.55) & 7.50\,\% (+0.31)\\
\bottomrule
\end{tabular}
\end{table}

\noindent
The HU\,FT preserves clean-Hungarian performance to within eval noise (the run-to-run variation in the CER metric) while blowing up dysarthric CER by 23.55\,pp absolute. This apparent selectivity is partly an artefact of pool composition: CV\_Hungarian was itself part of the fine-tuning mix, so its stability is not independent evidence of robustness --- when clean Hungarian is instead held out of training, it collapses too (Section~\ref{sec:dys-only}, Table~\ref{tab:pool-ablation}). We
interpret this as an adaptation paradox at the language-adaptation step:
a clean-data fine-tune on an LLM-decoder ASR damages dysarthric
distributions that the broader base pretraining had implicitly
handled. Whisper's larger
weakly-supervised pretraining buffer \citep{radford2023whisper}
gives it resistance; Qwen3-ASR-1.7B's tighter curated pretraining
does not. Accordingly, for the Qwen3 configurations we use the
\textbf{multilingual production Qwen3-ASR-1.7B checkpoint trained on
a 10-language pool including TORGO, UASPEECH, EasyCall and other
dysarthric corpora} (CER 2.91\,\% / WER 5.49\,\% on the held-out
multilingual eval) as the warm base, \emph{not} the corrupted
Qwen3-1.7B + HU\,FT.

\subsection{Dys-only pool follow-up}
\label{sec:dys-only}

To test whether the regression is driven by healthy-control content
specifically, we re-ran the HU\,FT on Hungarian\_Dysarthria only
(a 10K-utterance command corpus of Parkinson's-disease patients collected at the Budapest University of Technology and Economics; no healthy-control recordings). The regression got worse,
not better:

\begin{table}[t]\centering\scriptsize
\caption{Pool-composition ablation for the Qwen3 HU\,FT model. $\Delta$ denotes the worst-checkpoint change vs the Qwen3-ASR-1.7B
zero-shot reference (S1 43.61\,\%, CV\_HU 7.19\,\%).
Dys-only training breaks both held-out sets.}
\label{tab:pool-ablation}
\begin{tabular}{lrr}
\toprule
Pool & S1 $\Delta$ CER (pp) & CV\_HU $\Delta$ CER (pp)\\
\midrule
v1: HC-dominated 4-corpus & +23.55 (final) & +0.31 (final, preserved)\\
\textbf{v2: dys-only Hungarian\_Dys} & \textbf{+28.75 (ckpt75)} & \textbf{+22.55 (ckpt75)}\\
\bottomrule
\end{tabular}
\end{table}

S1 CER climbed +28.75\,pp at the 75\,\% checkpoint (still
rising), and CV\_HU also collapsed (+22.55\,pp) because
nothing in the training pool looks like clean Hungarian read speech.
v1 had preserved the performance on CV\_HU by coincidence --- CV\_Hungarian was inside
the v1 training mix. v2 over-fits to a 47-speaker
command-style PD corpus that is too narrow to either preserve
clean-Hungarian read speech or transfer to S1's
post-stroke read sentences. The mechanism is therefore \emph{not}
``healthy-control content is bad'' but \emph{``narrow language fine-tunes on Qwen3-1.7B were fragile in both pools tested''} --- a
corollary of finding\,7 in Section~\ref{sec:discussion}.

\subsection{Enrollment-time scaling}
\label{sec:enrollment-grid}

\begin{figure}[t]
\centering
\includegraphics[width=\linewidth]{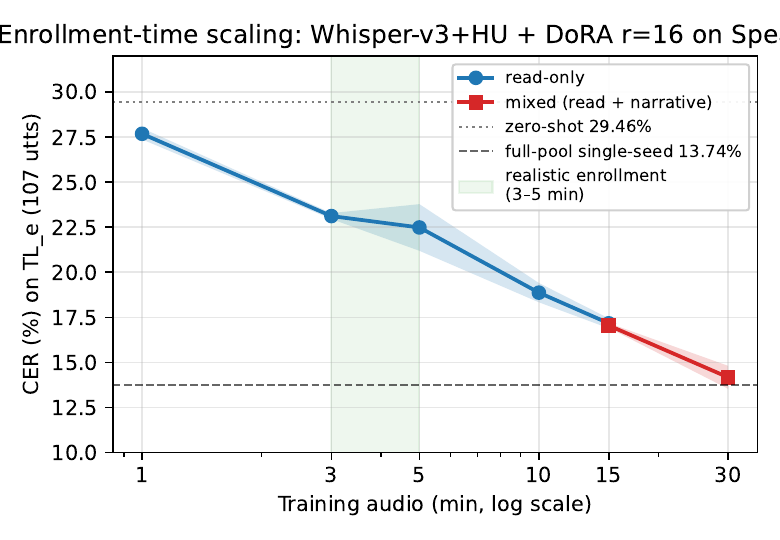}
\caption{CER vs train-pool minutes for DoRA $r{=}16$ on
Whisper-large-v3 + HU\,FT, three seeds per grid point. About 45.6\,\% of the
zero-shot-to-30-min CER reduction is reached within the first $\sim$5\,min;
performance keeps improving after 10 and even 30\,min.}
\label{fig:enrollment}
\end{figure}

The full-pool result (32.68\,min train audio $\to$ 13.74\,\% CER)
answers a research question but not the clinical one: \emph{how
little enrollment audio is enough?} We characterise this on the
per-patient recipe (Whisper-large-v3 + HU\,FT + DoRA $r{=}16$, which
LoRA matches; Section~\ref{sec:multiseed}) by
subsampling S1's train pool by accumulated audio-duration to
target wall-clock minute budgets of
$\{1, 3, 5, 10, 15, 30\}$\,min, with three random seeds per grid
point. For grid points $\le 21$\,min we draw read-sentence utterances
only (content-matched to the test split); for 30\,min we augment the data with
narrative (matches the full-pool composition); 15\,min is reported in
both compositions as a controlled comparison
(Table~\ref{tab:enrollment}, Figure~\ref{fig:enrollment}).

\begin{table}[t]\centering\scriptsize
\caption{Enrollment-time grid for DoRA $r{=}16$ on Whisper-large-v3
+ HU\,FT. Three seeds per configuration; mean $\pm$ standard deviation across
seeds. Zero-shot reference: 29.46\,\% CER.}
\label{tab:enrollment}
\begin{tabular}{rlrr}
\toprule
Min & Pool & CER (\%) & WER (\%)\\
\midrule
\phantom{0}1  & read     & 27.68 $\pm$ 0.31 & 47.36 $\pm$ 0.22\\
\phantom{0}3  & read     & 23.12 $\pm$ 0.17 & 43.99 $\pm$ 0.44\\
\phantom{0}5  & read     & 22.49 $\pm$ 1.29 & 41.55 $\pm$ 0.36\\
10  & read     & 18.87 $\pm$ 0.54 & 36.88 $\pm$ 0.78\\
15  & read     & 17.17 $\pm$ 0.28 & 34.35 $\pm$ 0.47\\
15  & mixed    & 17.05 $\pm$ 0.13 & 33.57 $\pm$ 0.11\\
30  & mixed    & 14.17 $\pm$ 0.64 & 30.01 $\pm$ 0.78\\
\bottomrule
\end{tabular}
\end{table}

\noindent\textbf{Findings from the enrollment grid.}
\textbf{(i)} Clinically realistic 3--5\,min enrollment yields
22--23\,\% CER --- a $\sim$7\,pp absolute / $\sim$24\,\% relative
reduction from zero-shot, which is 45.6\,\% of the reduction reached by
30\,min. \textbf{(ii)} Gains continue well beyond the first few minutes:
CER falls to 18.87\,\% at 10\,min and 14.17\,\% at 30\,min, so more than
half of the total zero-shot-to-30-min reduction accrues after the 5-min
point (tripling 10\,$\to$\,30\,min buys a further $\sim$4.7\,pp). \textbf{(iii)} Content composition
has zero effect at 15\,min: read-only 17.17\,\% vs mixed 17.05\,\%
is inside the seed-noise band. \textbf{(iv)} Seed stability is
tight (maximal standard deviation is 1.29 at 5\,min, in all other cases it is $\le$0.78). The 30-min
mixed three-seed mean (14.17\,\%) matches the full-pool single-seed
Section~\ref{sec:results} result (13.74\,\%) within 0.5\,pp.

\subsection{Same-base comparison: LoRA vs full and decoder-only
fine-tuning}
\label{sec:fullft-compare}

\begin{table}[t]\centering\scriptsize
\caption{Same-base comparison on Whisper-large-v3\,+\,HU\,FT (107-utterance
test). LoRA and zero-shot are from Tables~\ref{tab:main}--\ref{tab:multiseed};
full and decoder-only fine-tuning are three seeds (42/43/44,
mean\,$\pm$\,sd) with learning rates tuned on the 40-utterance validation
split. Serialised size is the bf16 artifact written per patient.}
\label{tab:samebase}
\resizebox{\linewidth}{!}{%
\begin{tabular}{lrrrrr}
\toprule
Method & CER (\%) & WER (\%) & Trainable & Serial.\ (MB) & Train (min)\\
\midrule
Zero-shot             & 29.46 & 49.53 & 0        & --   & --\\
Decoder-only full FT  & 22.96\,$\pm$\,0.08 & 49.70\,$\pm$\,0.75 & 906\,M & 3087 & 4.2\\
LoRA $r{=}16$         & 13.86\,$\pm$\,0.07 & 29.24\,$\pm$\,0.18 & 15.7\,M & 63 & 6.7\\
Full fine-tune        & 11.43\,$\pm$\,0.50 & 27.55\,$\pm$\,0.94 & 1.54\,B & 3087 & 7.6\\
\bottomrule
\end{tabular}
}
\end{table}

To weigh per-patient LoRA against the two alternatives it is usually
measured against --- full fine-tuning and full-parameter partial
fine-tuning --- we ran both on the identical base
(Whisper-large-v3\,+\,HU\,FT), the identical 262-utterance pool, and the
identical greedy \texttt{norm\_hu} scoring pipeline, three seeds each, with
learning rates tuned on the 40-utterance validation split
(Table~\ref{tab:samebase}). Full fine-tuning of all 1.54\,B parameters is the
most accurate adaptation of this base, giving 11.43\,\%\,$\pm$\,0.50 CER,
2.43\,pp below LoRA's 13.86\,\%, but writes a 3.1\,GB checkpoint file per patient compared to LoRA's 63\,MB adapter. Full-parameter fine-tuning of the decoder
with the encoder frozen is far worse (22.96\,\%\,$\pm$\,0.08 CER, its word
error rate essentially unchanged from the 29.46\,\%\,/\,49.53\,\% zero-shot
despite training 906\,M parameters), which already signals that the encoder
--- not raw parameter count --- is what adaptation needs here.

To locate where LoRA's residual gap to full fine-tuning lies, we ran an
adaptation-surface ladder on the same base and recipe, changing only the
LoRA target set (Table~\ref{tab:ladder}). Restricting LoRA to the decoder
attention alone --- freezing the encoder --- collapses to
24.21\,\%\,$\pm$\,0.06 CER, no better than the encoder-frozen decoder full
fine-tune (22.96\,\%): encoder-attention adaptation is the largest single contributor, and it is precisely the
encoder self-attention that the standard LoRA configuration
(Table~\ref{tab:main}) already adapts, recovering 10.15\,pp to
14.05\,\%\,$\pm$\,0.09. Adding the feed-forward projections
(\texttt{fc1}/\texttt{fc2}) to the LoRA target set recovers a further
1.96\,pp to 12.09\,\%\,$\pm$\,0.34 --- within 0.66\,pp of full fine-tuning
(11.43\,\%) --- at 28.8\,M parameters and 115\,MB, roughly 1.9\,\% of the
parameters and 3.7\,\% of the storage of the full model (WER 27.38 vs
27.55). The full-fine-tuning advantage is therefore a
feed-forward-plus-full-rank effect, not an encoder-attention effect: encoder
attention is the single largest LoRA lever but is already present in the
default configuration, and about three-quarters of the residual
full-fine-tuning gain is recoverable simply by extending the LoRA target set
to the feed-forward blocks, at minimal additional storage. Parameter-efficient
adaptation thus reaches near-full-fine-tuning accuracy at a small fraction of
the per-patient storage: the 63\,MB default trades 2.43\,pp for a
$\approx$49$\times$ storage reduction, and the 115\,MB feed-forward-augmented
variant trades only 0.66\,pp for a $\approx$27$\times$ reduction. (Learning
rates for full and decoder-only fine-tuning were validation-tuned; the LoRA
cells use the paper's fixed $1\times10^{-4}$.)

\begin{table}[t]\centering\scriptsize
\caption{LoRA target-set attribution ladder on Whisper-large-v3\,+\,HU\,FT
(107-utterance test, three seeds, mean\,$\pm$\,sd, fixed $1\times10^{-4}$
recipe as in Table~\ref{tab:main}). Only the LoRA target modules change;
each row adds a module group to the one above. Full fine-tuning is shown as
an anchor. This ladder is a single, internally consistent set of runs
executed together so its deltas are directly comparable; the
``$+$\,Encoder attention'' row is its re-run of the main LoRA configuration
reported in Table~\ref{tab:multiseed} (13.86\,\%), and the 0.19\,pp
difference between the two is run-to-run variation within the seed spread.}
\label{tab:ladder}
\begin{tabular}{lrrrr}
\toprule
LoRA target set & Params & Adapter (MB) & CER (\%) & WER (\%)\\
\midrule
Decoder attention only        & 10.5\,M & 42  & 24.21\,$\pm$\,0.06 & 44.81\,$\pm$\,0.15\\
$+$ Encoder attention (default) & 15.7\,M & 63  & 14.05\,$\pm$\,0.09 & 29.50\,$\pm$\,0.10\\
$+$ Feed-forward (\texttt{fc1}/\texttt{fc2}) & 28.8\,M & 115 & 12.09\,$\pm$\,0.34 & 27.38\,$\pm$\,0.48\\
\midrule
Full fine-tune (anchor)       & 1.54\,B & 3087 & 11.43\,$\pm$\,0.50 & 27.55\,$\pm$\,0.94\\
\bottomrule
\end{tabular}
\end{table}

Because the validation and evaluation splits overlap exactly with
\citet{mihajlik2025tts} (Section~\ref{sec:method}), their results provide
historical context on the same speaker and 107-utterance evaluation set.
Using the 195 read training utterances, they report 8.9\,\% CER for full
fine-tuning of the monolingual FastConformer\_Hu and 14.7\,\% for full
fine-tuning of Whisper-large-v3-turbo (Whisper-turbo, in that study); their
best results with synthetic TTS augmentation are 7.3\,\% and 12.2\,\%
respectively. Our LoRA configuration reports 13.86\,\% CER using a larger
training pool (262 utterances including 67 narrative), a different Whisper
checkpoint with a Hungarian warm start, and a different optimisation
procedure. The reported value is numerically lower than their real-data-only
Whisper result, but the experiments are not controlled or directly
comparable and do not establish that parameter-efficient adaptation
outperforms full fine-tuning; our own same-base comparison
(Table~\ref{tab:samebase}) makes that controlled statement directly, and
agrees with theirs that full fine-tuning is the more accurate of the two.
Their FastConformer\_Hu result remains substantially lower than
ours; adapter adaptation of that checkpoint has not been tested, and the
NeMo experiments in Section~\ref{sec:nemo-footnote} concern different
backbones. Their FastConformer\_Hu checkpoint is not publicly released,
mirroring our own internal Qwen3-ASR production checkpoint: each study's
strongest non-Whisper system rests on a private model, so the two are
symmetric in reproducibility. Their synthetic-TTS route is a complementary
data-centric axis
whose interaction with per-patient adapters is untested. The trained LoRA
artifact is 63\,MB against a full per-patient model checkpoint under full
fine-tuning; a fleet-level total-storage analysis across many patients is not conducted here.

\section{Discussion}
\label{sec:discussion}

We list our seven findings and their production / clinical-deployment implications.

\textbf{1. The LoRA family wins; the rest fall short of it.}
In the screen (Table~\ref{tab:main}) LoRA, DoRA and the QLoRA
parameterisation cluster tightly while the other four lag well behind
(the within-family verdict is finding~2). Of the remaining four
variants, LoHA is the closest:
+18.6\,\% relative CER reduction on Whisper (29.46\,$\to$\,23.99\,\%
CER) and +11.8\,\% on Qwen3, so LoHA fails relative to the LoRA
family, not in absolute terms. AdaLoRA, VeRA and VB-LoRA all land
within $\sim$1\,pp of zero-shot on Whisper and within $\sim$3\,pp
on Qwen3, i.e.~they are not useful adaptation alternatives. The per-patient small-data
regime ($\sim$80 optimiser steps) is the binding constraint:
methods that need many steps to converge to an internal allocation or
scaling decision (AdaLoRA, VeRA, VB-LoRA) have too small a budget,
and the Hadamard reparameterisation (LoHA) does not recover the
speaker-dependent dysarthric distribution shift the way it recovers
a speaker-disjoint children's-ASR shift (as in \citep{ankita2026children}).

\textbf{2. LoRA and DoRA tie; we choose LoRA for future work; 4-bit QLoRA
underperforms.} The multi-seed head-to-head (Section~\ref{sec:multiseed},
Table~\ref{tab:multiseed}) settles the within-family question that the
single-seed screen could not. Across three seeds, LoRA and DoRA are
statistically indistinguishable on both bases (paired bootstrap
$p{=}0.79$ Whisper, $p{=}0.55$ Qwen3), and LoRA's mean is in fact
marginally lower on each --- DoRA's single-seed Whisper lead in
Table~\ref{tab:main} did not survive re-seeding. We therefore adopt
LoRA as the per-patient production lead: it matches DoRA in accuracy
while being simpler, carrying no magnitude-vector overhead, and
training in roughly half the Whisper wall-clock time (6.7 vs 12.7\,min).
DoRA remains an alternative with no detected difference for anyone who prefers its
magnitude/direction decomposition. Real 4-bit NF4 QLoRA is
consistently worse than both on every seed ($+$0.69\,pp Whisper,
$+$1.99\,pp Qwen3 CER) and, at this 1.5--1.7\,billion-parameter scale, saves no peak
VRAM, so we do not recommend 4-bit quantisation in this small-data
per-patient regime. We default our production per-patient template to
LoRA $r{=}16$ on both bases, pending a later multi-speaker validation.

\textbf{3. Storage-efficient variants do not transfer to the per-patient application.}
VeRA (5\,MB adapter on Whisper, 11\,MB on Qwen3) and VB-LoRA
(503\,MB on Whisper, 313\,MB on Qwen3) both land within $\sim$1\,pp of zero-shot on Whisper and under 4\,pp on Qwen3 --- effectively no adaptation happens. VB-LoRA's large serialised
adapters are a serialisation artefact, not a deployment footprint:
the size is dominated by dense per-module vector-selection logits
(125.8\,M parameters), while the shared vector bank has only 8{,}192 parameters; PEFT's default \verb|save_only_topk_weights=false| stores the
full logits, and setting it true collapses the adapter to a few MB. The per-patient
data budget is too small for VeRA's frozen-random-projection scheme
to converge to useful per-layer scales, and VB-LoRA's shared vector bank
does not transfer to single-speaker training.

\textbf{4. AdaLoRA needs more steps than a per-patient scenario gives.}
AdaLoRA's rank-reallocation budget converges to a near-zero update
within five epochs; final adapter is dense (94.6\,MB on Whisper,
58.8\,MB on Qwen3) but its quality is no better than that of the zero-shot solution.

\textbf{5. Warm-base choice transfers oppositely across bases.}
Hungarian fine-tuning of the base \emph{helps} Whisper (the
HU\,FT-merged Whisper-large-v3 zero-shots S1 at 29.46\,\% vs
$\sim$32.24\,\% for the raw v3) but \emph{hurts} Qwen3-ASR-1.7B
(+23.55\,pp regression on S1, Section~\ref{sec:qwen-regression}). The
choice of warm base must be made per-base; reusing a clean-language
fine-tune across architectures is not safe.

\textbf{6. Ankita's LoHA result does not transfer.}
\citet{ankita2026children} report LoHA as the best PEFT variant for
children's ASR on Whisper. Our LoHA configuration is 10.07\,pp worse than
LoRA on the same Whisper-v3 base in our speaker-dependent dysarthric
condition. The closest mechanistic interpretation is that LoHA's
effective-rank boost helps speaker-disjoint distribution shift
where many speakers must be modelled simultaneously, but is a
burden in the speaker-dependent regime where there is one distribution shift and only 80 steps in which to perform it.

\textbf{7. Pretraining-mix breadth appears to matter more than the actual architecture --- under the recipes tested.}
The NeMo Section~\ref{sec:nemo-footnote} negative result + the Qwen3
HU\,FT regression Section~\ref{sec:qwen-regression} cohere as a single
tentative reading: for dysarthric PEFT, the breadth of the backbone's
pretraining mix may matter more than its architecture or its FLEURS /
Common Voice score on clean speech. Whisper-large-v3's 5-million-hour in-the-wild mix (the noisily-labelled audio used for its weakly-supervised pretraining) has a productive adaptation surface
for atypical speech under our recipe; the Granary-trained NeMo
backbones and Qwen3-ASR-1.7B did not adapt, under the same simple recipes.
We frame this as a recipe-level observation, not a structural claim:
we have not tried deeper adapter targets or higher ranks on those
backbones. The cautious corollary for production base selection is
that clean-speech leaderboards are not reliable indicators of
dysarthric-PEFT readiness.

\textbf{Clinical deployment: enrollment audio.}
The Section~\ref{sec:enrollment-grid} enrollment-time sweep answers the practical clinical question of how much patient audio a clinic needs to collect. A 3--5\,min recording session reduces CER from the 29.46\,\% zero-shot baseline to $\sim$22--23\,\%, which is 45.6\,\% of the reduction reached by 30\,min. Further enrollment continues to help substantially --- 18.87\,\% at 10\,min and 14.17\,\% at 30\,min --- so more than half of the total gain accrues after the first few minutes, and 30\,min is the largest condition tested rather than a plateau. For per-patient deployment we therefore treat a $\sim$3--5\,min session as a useful minimum rather than a sweet spot: it delivers a substantial CER reduction quickly, and more audio materially helps where the patient can provide it. \emph{Caveat:} this enrollment curve is
measured on a single speaker, in a single language, with severe
post-stroke dysarthria. The shape of the curve at other severities and
aetiologies is a
question we defer to follow-up work (Section~\ref{sec:conclusion}).

\textbf{Limitations.} This study rests on a single speaker of a single language with severe dysarthria, and uses a fixed five-epoch training budget with greedy decoding. We apply no content-mix balance control at the short-data end beyond the 15-min read-vs-mixed contrast. Multi-patient generalisation and the
FiLM-vs-LoRA fleet-Pareto question are deferred to future work
(see Section~\ref{sec:conclusion}).

\section{Practical recipe and release}
\label{sec:practical-recipe}

The deliverable accompanying this paper is a small standalone repository
(\texttt{per-patient-peft}) that bundles:

\begin{itemize}\setlength{\itemsep}{0pt}
  \item Separate training recipes for each variant (drop-in
        replacement for plain-LoRA per-patient pipelines, with the
        LoRA $r{=}16$ recipe as the default entry point on both
        Whisper-large-v3 and the Qwen3-ASR production checkpoint).
  \item A patient-side runtime that loads the shared base plus the small
        per-patient adapter and transcribes audio; the base weights
        (including the bulk of the audio encoder) are cached once and shared
        across patients, with only the low-rank adapter stored per patient.
  \item A simple ``adapter manager'' for the multi-patient case
        (one base, many adapters, switchable at inference time).
  \item The locked PEFT-variant Pareto chart and summary table,
        plus the enrollment-grid script.
\end{itemize}

\textbf{What will be released (source-available, on publication):} All
scripts, recipes, training configurations, the Pareto plotting script and
the enrollment-grid runner, under a research-use licence (commercial use
reserved pending a separate commercial track). We describe this as
source-available rather than open-source, since the licence restricts
field of use.

\textbf{What is restricted:} The S1 corpus, all adapter weights
trained on it, and the per-utterance result JSON files (which embed
verbatim reference transcripts) are covered by the signed Data
Sharing Agreement with the data subject, which allows research use and
permits onward transfer only with the data subject's written consent; we
therefore do not redistribute them openly. Adapter weights and per-utterance
JSON files are not part of the open release, but are available to bona fide
researchers on request, subject to the data subject's written consent and a
research-use DSA mirroring the original. Aggregate summary tables, Pareto
coordinates, and the main results in this paper are public.

\textbf{Choosing the production variant.} LoRA is our default on both
bases. The multi-seed head-to-head (Table~\ref{tab:multiseed}) shows LoRA and DoRA are statistically tied, so we take the simpler, cheaper of
the two; DoRA can be a drop-in alternative for anyone who prefers it. We
do \emph{not} recommend 4-bit QLoRA in this regime: it is worse on
every seed and both bases and saves no memory at this model scale.

\section{Conclusion and future work}
\label{sec:conclusion}

We have presented a single-speaker production case study --- the
first speaker-dependent, per-patient PEFT comparison on dysarthric
speech across seven LoRA-family variants and two production ASR
bases. A single-seed comparison isolated the LoRA family (LoRA, DoRA,
QLoRA) from the other four variants; a three-seed head-to-head with
\emph{real} 4-bit NF4 QLoRA then detected no significant LoRA--DoRA
difference on either base (paired bootstrap $p>0.5$; Whisper
13.86\,\%\,/\,13.90\,\% CER, Qwen3 28.10\,\%\,/\,28.33\,\%), while
true 4-bit QLoRA proved worse on every seed (14.56\,\% / 30.09\,\%) with
no memory saving at this scale. We prefer the simpler, cheaper LoRA as
the per-patient production lead, with DoRA an alternative with no detected difference.
The other four variants (LoHA, AdaLoRA, VeRA, VB-LoRA) did not reach the LoRA family within the $\sim$80-optimiser-step budget the
per-patient regime allows, though LoHA still delivers an 18.6\,\%
relative CER reduction on Whisper. A 6-point enrollment-time grid on
the production recipe shows $\sim$5\,min of patient audio captures 45.6\,\% of the zero-shot-to-30-min CER reduction (a $\sim$7\,pp CER reduction), with substantial further gains to 30\,min, on a single severe post-stroke speaker. We caution that all
fleet-level recommendations here rest on just one speaker and await the
multi-patient validation described below.

Future work will extend the comparison to a multi-patient cohort
spanning aetiologies, enabling a proper benchmark against FiLM
speaker-conditioning \citep{lopez2026film} and a fleet-level Pareto test over storage, quality, and multi-task retention. A definitive
multi-LoRA-recipe sweep on the NeMo backbones
(Section~\ref{sec:nemo-footnote}) --- deeper LoRA targets, higher-rank
LoHA / DoRA --- is also queued as a follow-up: under the simple
recipes we tested, both NeMo backbones regress, but the structural
interpretation is not yet definitive.

\section*{Acknowledgements}
We thank the speaker, S1, for contributing the corpus under a research-use
Data Sharing Agreement, and the host institution for hosting the data
collection.

\section*{Ethics and data statement}
This study uses recordings that the single participant (S1, a pseudonym)
made of his own voice and contributed for this research under written
informed consent and a signed Data Sharing Agreement (research use only; no
redistribution of the audio or its derivatives), including their reuse in
this study and in the related work of \citet{mihajlik2025tts}. Because the
study analyses one individual's own, self-recorded speech, with that
individual's informed consent and no other human participants and no
clinical intervention, it did not require institutional ethics-committee
approval, and none was sought. Because we disclose that S1 is the same
participant as in \citet{mihajlik2025tts}, the de-identification is
pseudonymous rather than anonymous --- a determined reader could infer the
participant's identity --- so we do not describe the data as anonymised. We
use the data solely for the stated research purpose, make no attempt at
re-identification, and process it in compliance with the UK Data Protection
Act 2018 / UK GDPR; the trained adapters and per-utterance outputs are
withheld under the same agreement (Section~\ref{sec:practical-recipe}).

\section*{Declaration of generative AI and AI-assisted technologies in the manuscript preparation process}
The first author communicates exclusively via AI-assisted eye-gaze
interface due to motor neurone disease. Generative AI (Claude Opus 5,
Anthropic) was used as an assistive communication tool throughout the
research process: experimental design, code development, data
analysis, statistical computation, figure generation, and manuscript
drafting. All scientific decisions, interpretations, and conclusions
were made by the authors. The AI served as an accessibility tool
enabling a researcher with severe physical disability to conduct
computational research --- analogous to a screen reader for visually
impaired researchers. No AI-generated content was presented without
author review and verification. After using this tool/service, the
author(s) reviewed and edited the content as needed and take(s) full
responsibility for the content of the published article. AI use in the
research process is described in Section~\ref{sec:ai-research}.

\section*{CRediT authorship contribution statement}
\textbf{Bernard Muller:} Conceptualization, Methodology, Software,
Formal analysis, Investigation, Visualization, Writing -- original
draft. \textbf{L\'aszl\'o T\'oth:} Resources, Data curation,
Validation, Writing -- review \& editing. \textbf{LaVonne Roberts:}
Project administration, Supervision, Writing -- review \& editing.

\section*{Declaration of competing interest}
The authors declare that they have no known competing financial
interests or personal relationships that could have appeared to
influence the work reported in this paper. The Scott-Morgan Foundation develops the per-patient dysarthric-ASR system described here and reserves commercial rights on the released code under the research-use licence; no author receives personal income from it.

\section*{Data availability}
The S1 corpus, the adapters trained on it, and the per-utterance
outputs are restricted under a signed Data Sharing Agreement and are not
openly redistributed (Section~\ref{sec:practical-recipe}); they are
available to bona fide researchers on request, subject to the data subject's
written consent and a research-use agreement mirroring the original. The training scripts, per-variant configurations, and the
Pareto / enrollment runners will be released, source-available under a
research-use licence, on publication. Aggregate summary tables and the
numbers reported in this paper are public.

\section*{Funding}
This work was carried out as part of the research programme of The
Scott-Morgan Foundation; it received no specific grant from public,
commercial, or not-for-profit funding agencies.

\bibliographystyle{elsarticle-num-names}
\bibliography{refs}

\end{document}